# Towards A Deep Insight into Landmark-based Visual Place Recognition: Methodology and Practice

Bo Yang[1, 2], Jun Li[2, 3], Xiaosu Xu[1], Hong Zhang[2]

[1] Key Laboratory of Micro-Inertial Instrument and Advanced Navigation Technology, Ministry of Education, School of Instrument Science and Engineering, Southeast University, Nanjing, China.

[2] Department of Computing Science, University of Alberta, Edmonton, Alberta, Canada.

[3] School of Automation, Southeast University, Nanjing, China.

**Abstract** In this paper, we address the problem of landmark-based visual place recognition. In the state-of-the-art method, accurate object proposal algorithms are first leveraged for generating a set of local regions containing particular landmarks with high confidence. Then, these candidate regions are represented by deep features and pairwise matching is performed in an exhaustive manner for the similarity measure. Despite its success, conventional object proposal methods usually produce massive landmark-dependent image patches exhibiting significant distribution variance in scale and overlap. As a result, the inconsistency in landmark distributions tends to produce biased similarity between pairwise images yielding the suboptimal performance. In order to gain an insight into the landmark-based place recognition scheme, we conduct a comprehensive study in which the influence of landmark scales and the proportion of overlap on the recognition performance is explored. More specifically, we thoroughly study the exhaustive search based landmark matching mechanism, and thus derive three-fold important observations in terms of the beneficial effect of specific landmark generation strategies. Inspired by the above observations, a simple yet effective dense sampling based scheme is presented for accurate place recognition in this paper. Different from the conventional object proposal strategy, we generate local landmarks of multiple scales with uniform distribution from entire image by dense sampling, and subsequently perform multi-scale fusion on the densely sampled landmarks for similarity measure. The experimental results on three challenging datasets demonstrate that the recognition performance can be significantly improved by our efficient method in which the landmarks are appropriately produced for accurate pairwise matching.

## 1 Introduction

Visual place recognition, which is widely used in the localization and navigation systems [1-7], aims to identify whether the current view corresponds to the place or location that has been already visited [8]. Although considerable progress has been made in this field, a wide range of variances pose great challenges to the landmark matching, and thus severely degrade the recognition performance. For instance, the appearance of a location varies drastically with significant illumination and seasonal changes in dynamic environments. Besides, it is difficult to guarantee a place can be revisited from the same viewpoint and position as before [8]. In general, the environment and viewpoint variations are two major challenges in place recognition, and thus considerably deteriorate the recognition performance [8, 1, 4]. Recently, a landmark-based visual place recognition method is proposed to address the above two issues simultaneously [9]. To be specific, a set of candidate regions containing particular landmarks are first extracted by exploiting object proposal methods, and represented by CNN feature produced from an off-the-shelf pre-trained deep model. Subsequently, the pairwise similarity matching between different CNN features in two images is performed in an exhaustive manner and accumulated by pooling strategy for similarity evaluation. Although this method somewhat enables handling both environment and viewpoint changes without any training process involved, it produces massive landmark-dependent image patches exhibiting significant distribution variance in scale and overlap, and consequently results in biased similarity between pairwise images yielding the suboptimal performance due to the inconsistency in landmark distributions. More specifically, this approach suffers from the following two limitations:

**1. Relatively small-size landmarks have a higher mismatch probability.** Fig. 1 gives an example illustrating the mismatched landmarks achieved by conventional landmark-based visual place recognition method. It can be readily observed that most incorrect matches have relatively small size. Intuitively, their area ratios, which is defined as the landmark area over the full image area, generally account for less than 10%. Furthermore, these small-size outliers exist between all the pairwise images and play a dominant role in similarity measure. As a result, this leads to the biased results and degraded recognition accuracy.

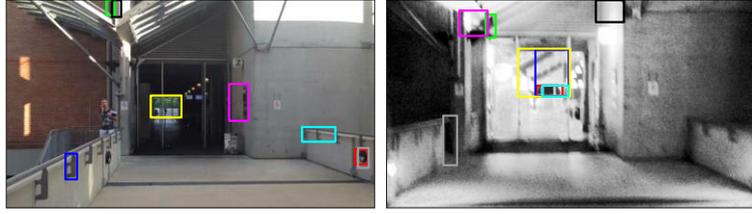

Fig. 1. An example of mismatched landmarks between two images on the Gardens Point Campus dataset. The matched pairwise landmarks are highlighted in boxes of the same color. It is shown that incorrect matches have relatively small size.

**2. There exist highly overlapped matched landmarks leading to the redundancy in similarity matching.** Fig. 2 illustrates an example indicating the overlapped matched results between two irrelevant images. The two pairwise matched results are outlined in red and green boxes respectively. Matching the pairwise landmarks with high Intersection of Union (IoU) essentially operates by repeatedly calculating the two regions of close neighborhood, and thus is likely to result in redundant similarity matching.

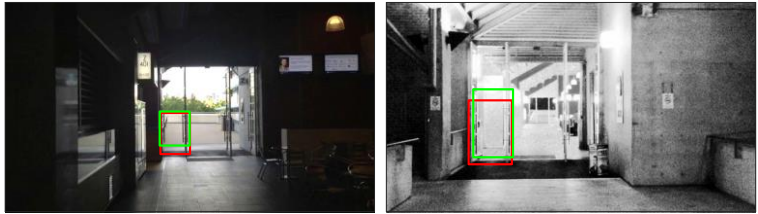

Fig. 2. An example of highly overlapping matched landmarks between two irrelevant images on the Gardens Point Campus dataset. The matched pairwise landmarks are highlighted in boxes of the same color.

In this paper, we make a comprehensive study in which the impact of landmark scale and overlap on the recognition performance is thoroughly explored. Furthermore, we develop a dense sampling based method to overcome these limitations for improving the recognition performance. More specifically, we make a two-fold in-depth evaluations addressing the above-mentioned limitations. On the one hand, inspired from the study of feature matching repeatability with respect to scale [10], we study the three characteristics of matched landmarks corresponding to the first limitation with respect to landmark scale. In addition, we design two landmark selection methods to reduce the negative influence of the small-size landmarks in a comparative study. On the other hand, inspired by the non-maximum suppression (NMS) [11] and soft-NMS methods [12] in object detection, we design two methods to suppress the effect of the redundancy in matching.

In summary, we arrive at the following three observations: **1) Larger-scale landmarks generally work better than their smaller counterparts for pairwise matching in the exhaustive search. 2) The recognition rate is largely dependent on the distribution of landmarks with different scales. 3) The recognition performance tends to decline with a decrease in the proportion of overlap between pairwise landmarks, while desirable performance is achieved with a certain proportion.**

Motivated by these three observations, a simple yet effective scheme is presented for accurate place recognition in this paper. To be specific, we first impose dense sampling method on the entire image to generate landmarks of multiple scales instead of the conventional object proposal approach. Then, we perform multi-scale fusion on these landmarks for accurate place recognition. In particular, our method allows producing pre-calculated landmark positions, and thus it is more computationally efficient than the classic object proposal based approaches.

The primary contributions of our paper are summarized as follows:

1. We carry out a comprehensive study to thoroughly explore the impact of the landmark distribution on the landmark-based place recognition. Furthermore, we arrive at three important observations that benefit the practice in visual place recognition.

2. We develop a dense sampling based place recognition method for landmark-based place recognition. Compared with the traditional schemes, our method achieves promising recognition performance while enjoys desirable computational efficiency.

The rest of the paper is organized as follows. After reviewing the classical landmark-based place recognition method and introducing the challenging datasets as well as the evaluation procedure in Section 2, we deeply evaluate the influence of landmark scale and overlap on the performance in Section 3. In Section 4, we elaborate the dense sampling based method for landmark-based place recognition. Experiments are conducted in Section 5 before the final conclusions are drawn in Section 6.

## 2 Classical landmark-based visual place recognition review and datasets

In this section, we first briefly describe the processing pipeline of the classical landmark-based visual place recognition method [9], and then introduce the datasets that will be used for comprehensive study and performance evaluation.

### 2.1 Classical landmark-based visual place recognition

Classical landmark-based visual place recognition works in the following procedure: First, 100 landmarks are extracted from an image by the state-of-the-art object proposal method [13], and then described by a set of 64,896-dimensional ConvNet feature vectors built on the third convolutional layer (conv3) of the pre-trained AlexNet architecture [14]. Next, the dimension of each feature is reduced to 1,024 by Gaussian Random Projection method for compact representation. Subsequently, the landmark matching is performed by adopting the nearest neighbor search based on the cosine distance of 1,024 dimensional features and only reciprocal matches are identified as true matches. Finally, the overall similarity between two images can be calculated as [9]:

$$S = \frac{1}{\sqrt{n_a * n_b}} \sum_{ij} 1 - (d_{i,j} * s_{i,j}) \tag{1}$$

where $d_{i,j}$ is the cosine distance between two matched landmarks; $n_a$, $n_b$ are the number of extracted landmarks proposals in both images, including the non-matched ones (generally, both equal to 100) and $s_{i,j}$ is defined as shape similarity calculated as [9]:

$$s_{i,j} = \exp\left(\frac{1}{2}\left(\frac{|w_i - w_j|}{\max(w_i, w_j)} + \frac{|h_i - h_j|}{\max(h_i, h_j)}\right)\right) \tag{2}$$

In summary, the process of this approach is briefly illustrated in Fig. 3.

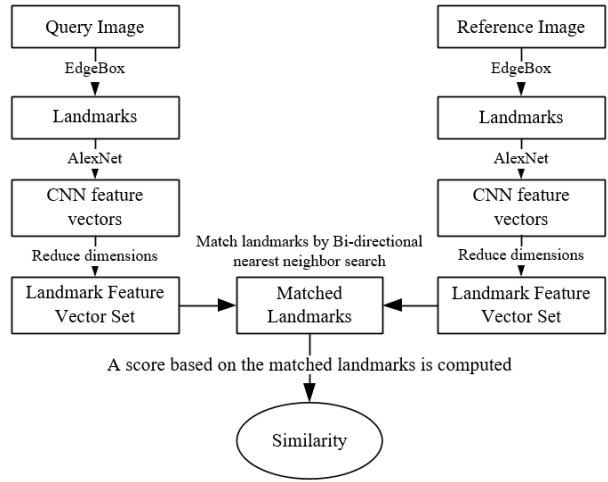

Fig. 3. The process of the classical landmark-based visual place recognition approach

In [15], it is shown that the ConvNet feature from the layer conv3 of the AlexNet enjoys preferable invariance to the environment variations. Meanwhile, describing a scene by a series of landmark regions significantly improves the robustness to viewpoint changes [9]. Thus, the landmark-based visual place recognition method reveals superior recognition performance in the challenging situations when both environment and viewpoint variations are present. In addition, according to [16], the state-of-the-art performance is achieved by leveraging Edge Box for extracting landmarks and pre-trained AlexNet model for landmark representation. Therefore, they are involved in the baseline place recognition method in this paper.

### 2.2 Datasets and evaluation procedure

Three public benchmark datasets are used in this paper for visual place recognition, namely Gardens Point Campus dataset [17], Mapillary Berlin [18] and Freiburg Across Seasons [6]. In particular, Gardens Point Campus dataset is used for the empirical study while the evaluations are carried out on the remaining two challenging datasets. For performance measure, we use precision-recall curves for evaluation metric.

**Gardens Point Campus** [17] includes three sequences named as 'day-left', 'day-right' and 'night-right' respectively recoded on the Gardens Point Campus, Queensland University of Technology. With the same loop involved, the first two sequences are collected during the day on the both left and right side of the path, while the third sequence is recorded on the right side during the night. In practice, the 'day-left' and 'night-right' sequences which manifest significant illumination and viewpoint variations are used as query and reference respectively for the comprehensive study in Section 3.

**Mapillary Berlin** [18] is a dataset assembled from the Mapillary which is a crowdsourced alternative to Google Street View. It includes 157 query images and 67 reference images which are all captured on the same street of Berlin during different time. This dataset contains large viewpoint changes, moderate environment variations along with dynamic objects, and thus poses great challenges to visual place recognition.

**Freiburg Across Seasons** [6] includes three sequences respectively named as 'Summer 2012', 'Summer 2015' and 'Winter 2012'. All three sequences are captured in Freiburg in May 2012, 2015 and December 2012 respectively. In this paper, we use the most challenging two sequences, 'Summer 2015' and 'Winter 2012' involving drastic illumination, viewpoint, and scene variations during three years. In addition, there exist multiple images corresponding to one place for one sequence in the original dataset. However, only one image is preserved in our experiments due to the evaluation approach mentioned below.

In this paper, we evaluate different methods by using precision-recall curves. In terms of the evaluation metric, the original similarity matrix encoding the image similarities between all possible image pairings is transformed to precision-recall curve by following the approach presented in [15]:

For performance measure, a predefined threshold, which is computed as the distance ratio of the second best match over the best match in the nearest neighbor search, is carefully determined to distinguish the positive and negative match in implementation. Besides, a true positive match is recognized when it is within adjacent frame of the ground truth (depending on the frame rate of the recorded dataset), whilst the remaining frames are considered as false positive. The tuning parameter for generating the precision-recall curve is the distance threshold which is between 0 and 1. Note that this approach lends itself to the cases when at most one ground truth is considered corresponding to a single query. In our scenario, only one reference image is captured for one place, and thus this method is suitable for real-world applications.

## 3 Landmark-based place recognition: a comprehensive study

In this section, we deeply evaluate the influence of the various landmark scales and the overlap proportions on the performance of visual place recognition and arrive at three important observations.

### 3.1 The impact of various landmark scales

In this subsection, we explore the influence of the various landmark scales. Analogously, a comprehensive study is conducted in [10] to explore the performance of detected features for guiding the feature extraction and selection process. To be specific, it presents a scale dependent feature selection method through learning repeatability statistics of SIFT features with respect to scale. Following this work, we design two landmark selection schemes based on studying the three characteristics of landmark features with respect to the scales:

**Correct landmark matching rate**, which is denoted as '**CMR**' for short, is defined as the ratio of true positive matched landmarks over whole matched landmarks in matching two query-dependent ground truth images.

**Contributions of landmarks with different scales to similarity evaluation**, which is denoted as '**CLS**' for short, is quantitatively computed as the ratio of similarity matching on the landmarks of individual scale over all scales in matching query-dependent ground truth images and query-irrelevant images.

**Average similarity for all matched landmarks between pairwise images**, which is denoted as '**ASL**' for short, is defined as the average similarity of matched landmarks with respect to different scales between pairwise images. This characteristic is also counted respectively in matching query-dependent ground truth images and query-irrelevant images.

To our knowledge, previous research fails to explore the impact of various landmark scales on the matching accuracy. Inspired by the beneficial effect of the multi-scale pooling method in the field of image matching [19], we select area ratio as the landmark scale in this paper:

$$area\ ratio\ of\ landmark = Landmark\ area\ /\ Full\ Image\ area \qquad (3)$$

It is shown that the area ratio defined in the above formulation is essentially the normalized scale distributed in the range of [0,1]. In the following, we term the area ratio as normalized scale.

Similar to the scale-space in SIFT features [20], we define nine different landmark scales from 'Scale 1' to 'Scale 9' at an interval of roughly $\sqrt{2}$ as shown in Table 1.

Table 1 Nine landmark scales with the corresponding range of normalized scale

| Scale Index | Scale 1 | Scale 2 | Scale 3 | Scale 4 | Scale 5 | Scale 6 | Scale 7 | Scale 8 | Scale 9 |
|---|---|---|---|---|---|---|---|---|---|
| normalized scale | 0~0.02 | 0.02~0.05 | 0.05~0.09 | 0.09~0.14 | 0.14~0.23 | 0.23~0.34 | 0.34~0.5 | 0.5~0.72 | 0.72~1 |

To illuminate these three characteristics of landmark features mentioned above with respective to scales, we respectively use 200 images from 'day-left' and 'night-right' sequences as query and reference images in Gardens Point Campus dataset and match extracted landmarks between query and reference images through the bi-directional nearest neighbor search. Furthermore, for these matched landmarks in two query-dependent ground truth images, we recognize and annotate the true and false label manually. Subsequently, we study the characteristics based on these matched landmarks and labels. Experimental results are shown in Fig. 4.

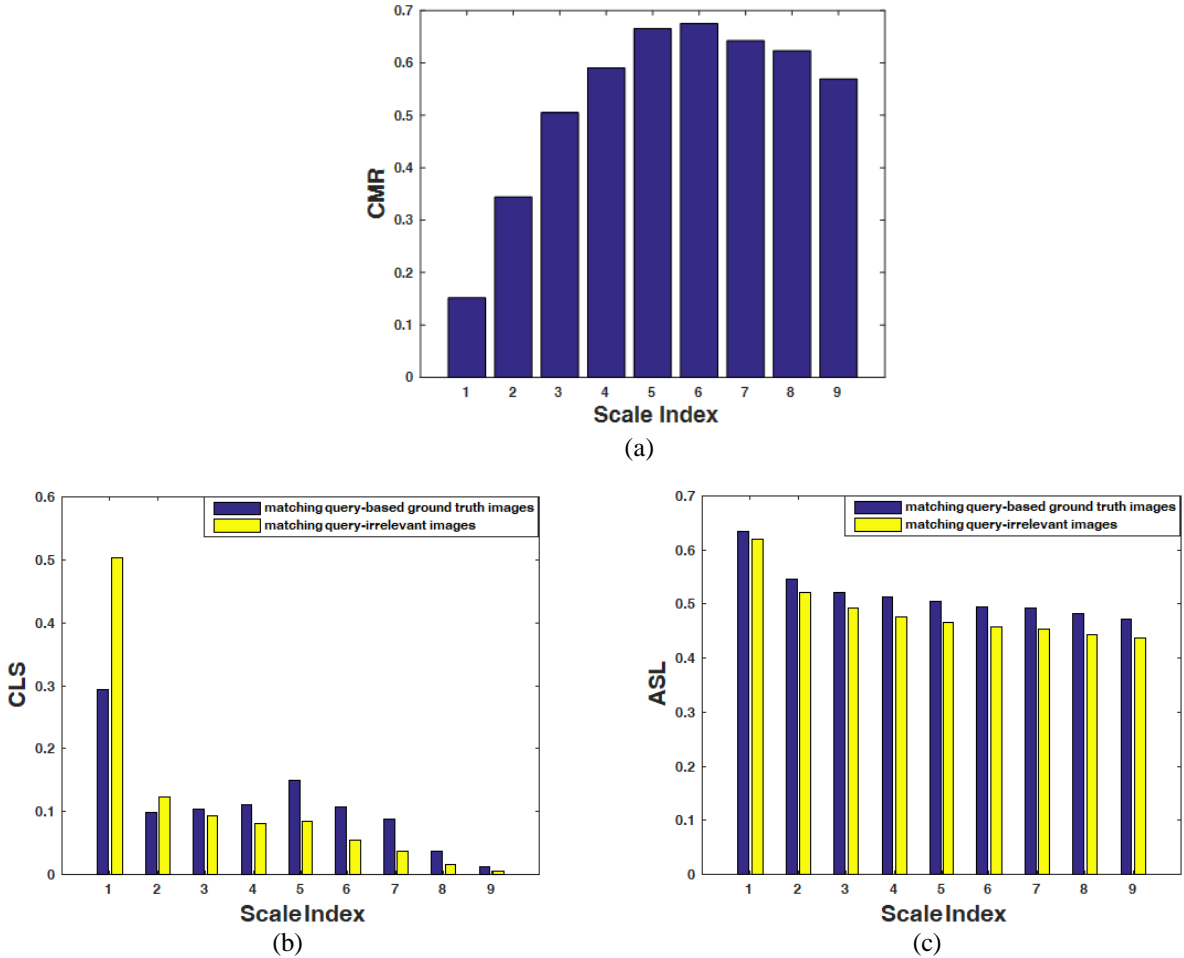

Fig. 4. (a) Correct landmark matching rate of various scales based on bi-directional nearest neighbor search between the 'day-left' and 'night-right' sequences. (b) Contributions of landmarks with different scales to similarity evaluation in matching query-dependent ground truth images and query-irrelevant images respectively. (c) Average similarity for all matched landmarks with respect to scales between pairwise images in matching query-dependent ground truth images and query-irrelevant images respectively.

In Fig. 4, it is clearly shown that larger-scale landmarks such as Scale 5, 6 and 7 lead to reliable matching results relatively. By contrast, landmarks in smaller-scale such as Scale 1, 2 and 3 have a lower correct matching rate, but these landmarks, particularly in Scale 1, have a higher average similarity and contribute the most of the scores to similarity evaluation, and thus lead to many mismatched landmark pairs and biased similarity in matching two images.

On the other hand, the landmarks with smaller scales, particularly those in Scale 1, contribute less to the similarity evaluation in matching query-specific ground truth images than matching query-irrelevant ones. By contrast, matching query-specific ground truth images outweighs query-irrelevant ones in similarity measure for larger landmarks. Besides, the difference between the average similarity in matching ground truth images and two query-irrelevant images for large-scale landmarks are greater than the ones with smaller scales. This implies the larger landmarks (from Scale 4 to 7) are capable of better distinguishing the correct matches from incorrect ones. In addition, since limited landmarks in Scale 8 and 9 are extracted, they hardly produce any influence on the recognition performance.

To summarize, we arrive at the first observation: larger-scale landmarks generally work better than their smaller counterparts for pairwise matching in the exhaustive search.

Motivated by this observation, we follow the feature selection method in [10] by designing two strategies to reduce the negative impact of smaller-scale landmarks. The pseudocodes of our method are shown in Algorithm 1 and 2. Specifically, for the sake of efficiency, a total number of 1000 candidate regions are extracted firstly by Edge Box per image for two designed schemes in practice. Subsequently, in Scheme 1, top 100 landmarks with scales larger than Scale 3 are selected from the set of candidate regions. In Scheme 2, by contrast, top 100 landmarks are selected in a scale-specific manner. Thereafter, these selected landmarks are delivered to the subsequent pairwise matching for place recognition, which is analogous to the classical method [9].

---

**Algorithm 1** Landmark Selection Scheme 1
**Input:** 1000 extracted landmarks by Edge Box $E$
**Output:** 100 selected landmarks $L$
  $i = 1$
  **while** number of selected landmarks $N < 100$ **do**
    **if** Scale of $i$-th extracted landmarks $> 3$
      $L \leftarrow E(i)$
      $N = N + 1$
      $i = i + 1$
    **else**
      $i = i + 1$
    **end if**
  **end while**

---

**Algorithm 2** Landmark Selection Scheme 2
**Input:** 1000 extracted landmarks by Edge Box $E$
**Output:** 100 selected landmarks $L$
  $i = 1, j = 1$
  $S = [5, 6, 7, 8, 9, 4]$
  **while** number of selected landmarks $N < 100$ **do**
    **if** Scale of $i$-th extracted landmarks $== S(j)$
      $L \leftarrow E(i)$
      $N = N + 1$
      $i = i + 1$
    **else**
      $i = i + 1$
    **end if**
    **if** $i == 1001$
      $j = j + 1$
      $i = 1$
    **end if**
  **end while**

---

With these two schemes, most of the landmarks in smaller-scale (from Scale 1 to 3) are discarded while larger-scale landmarks are selected for place recognition. We test these two schemes on 'day-left' vs 'night-right' sequences from Gardens Point Campus dataset. Results are shown in Fig. 5.

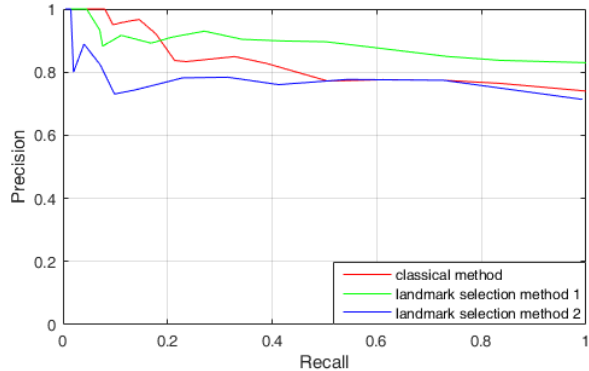

Fig. 5. Results of place recognition methods with two landmark selection schemes and the classical method [9] on 'day-left' vs 'night-right' sequences from Gardens Point Campus dataset

As shown in Fig. 5, the Scheme 1 achieves the best performance and provides a significant improvement over the baseline. However, the Scheme 2 lags behind the baseline reporting inferior results. This can be explained by the difference between these two schemes in the landmark selection strategy, which thus leads to the variance in the landmark distribution. In order to gain an insight into the difference between these two methods, we explore the distribution of the landmarks extracted by two schemes for both query and reference images with respect to different scales. The specific statistical results are shown in Fig. 6.

It is clearly demonstrated that Scheme 1 produces the landmarks at multiple scale levels with more uniform distribution compared with Scheme 2. Therefore, we arrive at the second observation: the recognition rate is largely dependent on the distribution of landmarks with different scales. More specifically, the multiple scales and the uniform distribution in different scale landmarks contribute to improving the recognition performance.

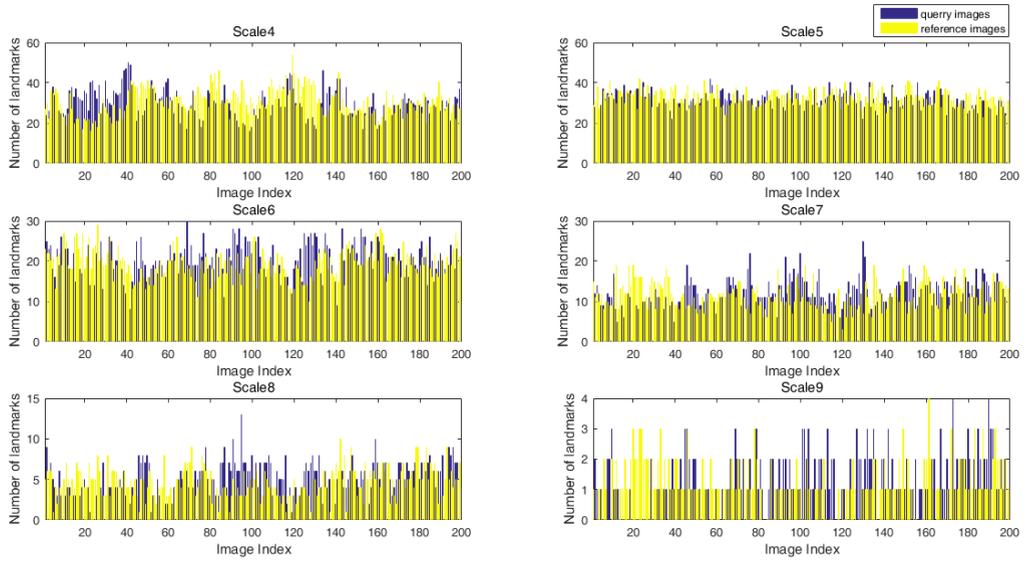

(a)

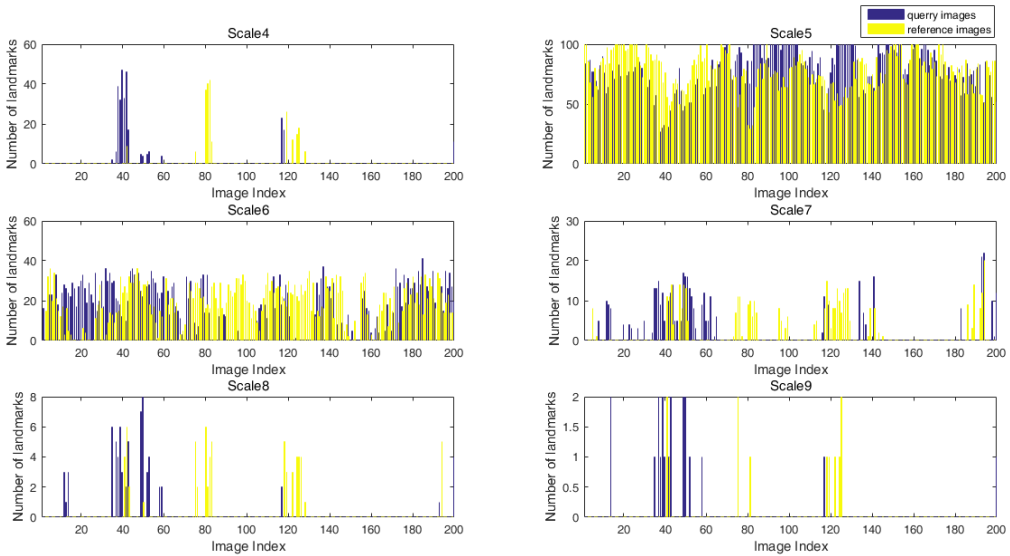

(b)

Fig. 6. (a) Distribution of landmarks selected by landmark selection Scheme 1 on 'day-left' and 'night-right' sequences in Gardens Point Campus dataset. (b) Distribution of landmarks selected by landmark selection Scheme 2 on 'day-left' and 'night-right' sequences in Gardens Point Campus dataset.

**3.2 The impact of overlap proportions**

In this subsection, inspired by the non-maximum suppression (NMS) [11] and soft-NMS methods [12] in the field of object detection, we design two overlap reduction strategies to explore the influence of the overlapping landmarks in images for visual place recognition. The pseudocodes of the schemes are shown in Algorithm 3 and 4, respectively. Besides, the Intersection over Union (IoU) is used for evaluating the proportion of overlap.

In the first scheme, we select 100 landmarks from 1000 candidates available per image with IoU lower than a specific threshold for visual place recognition. Thus, the influence of the overlapping landmarks can be reduced by discarding the highly overlapping landmarks. Thereafter, similar to the classical method, these selected landmarks are delivered to the subsequent pairwise matching for place recognition.

Different from scheme 1, the 100 landmarks extracted by Edge Box per image are also utilized in scheme 2, which is similar to the classical method. In terms of scheme 2, we use soft-NMS strategy to suppress the similarity scores between matched landmarks with high proportion of overlap, and then compute the similarity between two images based on the suppressed scores. Similar to the setup in [12], the parameter σ of the Gaussian penalty function is set to 0.5.

**Algorithm 3** Overlap Reduction Scheme 1
**Input:** 1000 extracted landmarks by Edge Box $E$
   Threshold of IoU $t$
**Output:** 100 selected landmarks $L$
   $i = 1$, $flag = 0$
   **while** number of selected landmarks $N < 100$ **do**
     **if** $N == 1$
       $L \leftarrow E(i)$
       $N = N + 1$
       $i = i + 1$
     **else**
       **for** $j = 1$ to $N$ **do**
         **if** IoU between $L(j)$ and $E(i) > t$
           $flag = 1$
         **end if**
       **end for**
       **if** $flag == 0$
         $L \leftarrow E(i)$
         $N = N + 1$
         $i = i + 1$
       **else**
         $i = i + 1$
         $flag = 0$
       **end if**
     **end if**
   **end while**

**Algorithm 4** Overlap Reduction Scheme 2
**Input:** Matched landmarks $M$
   Similarity scores between two matched landmarks $S$
   Threshold of IoU $t$
**Output:** Adjusted similarity scores between two matched landmarks $S$
   **while** $M \neq$ empty **do**
     $m \leftarrow \arg\max S$
     $M \leftarrow M - M(m)$
     **for** $i = 1$ to number of $M$ **do**
       **if** IoU between $M(i)$ and $M(m) > t$
         $S(i) \leftarrow S(i) * e^{-\frac{IoU(M(i),M(m))^2}{0.5}}$
       **end if**
     **end for**
   **end while**

These two schemes are evaluated on the 'day-left' vs 'night-right' sequences from Gardens Point Campus dataset. In scheme 1, the minimum IoU threshold is set to 0.4 in order to guarantee the number of landmarks selected from available landmarks reaches 100 per image, while its counterpart is set to 0.3 in scheme 2. Experimental results are shown in Fig. 7.

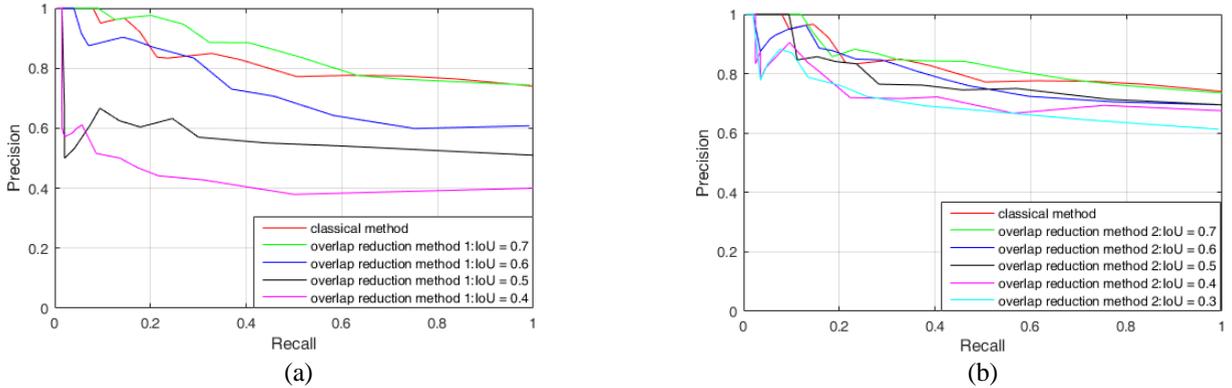

Fig. 7. (a) Results of the classical method and overlap reduction scheme 1 with IoU threshold from 0.4 to 0.7 on 'day-left' vs 'night-right' sequences from Gardens Point Campus dataset. (b) Results of the classical method and overlap reduction scheme 2 with IoU threshold from 0.3 to 0.7 on 'day-left' vs 'night-right' sequences from Gardens Point Campus dataset.

In Fig. 7, both schemes achieve the performance on par with the classical method when IoU threshold are set to 0.7. However, significant performance dropped is observed with the decrease in the IoU threshold. Thus, it hardly produces any performance improvement for place recognition by reducing or suppressing the impact of overlapping landmarks. Conversely, appropriate proportions of overlap benefit visual place recognition.

Although overlapping landmarks lead to redundancy in matching, it is also capable of enhancing the similarity matching between neighboring pairwise landmarks, and thus is somewhat robust to geometric variances. On the other hand, it sufficiently captures the visual cues in the entire image with favorable descriptive power.

To sum up, we arrive at the third observations: the recognition performance tends to decline with a decrease in the proportion of overlap between pairwise landmarks, while desirable performance is achieved with a certain proportion.

## 4 Practice: dense sampling based place recognition method

In the conventional landmark-based place recognition methods, the landmarks describing particular objects are usually obtained by object proposal methods, e.g. edge box [13]. This strategy mainly suffers from the following two limitations. On the one hand, it is observed from the above comprehensive study that the appropriate landmark distribution substantially affects the recognition performance. However, the landmarks extracted by object proposal method usually exhibit dramatic variances in the landmark scale and proportion of overlap changes. Thus, the object proposal based methods fail to appropriately capture the landmark distribution and cannot always guarantee the desirable recognition performance. On the other hand, the edge-box like object proposal methods are computationally expensive and thus are not suitable for the real-time applications. In this sense, inspired by the previous work in the field of image classification [21], we design a dense sampling based approach in which the distribution uniformity in the landmark scale are sufficiently characterized.

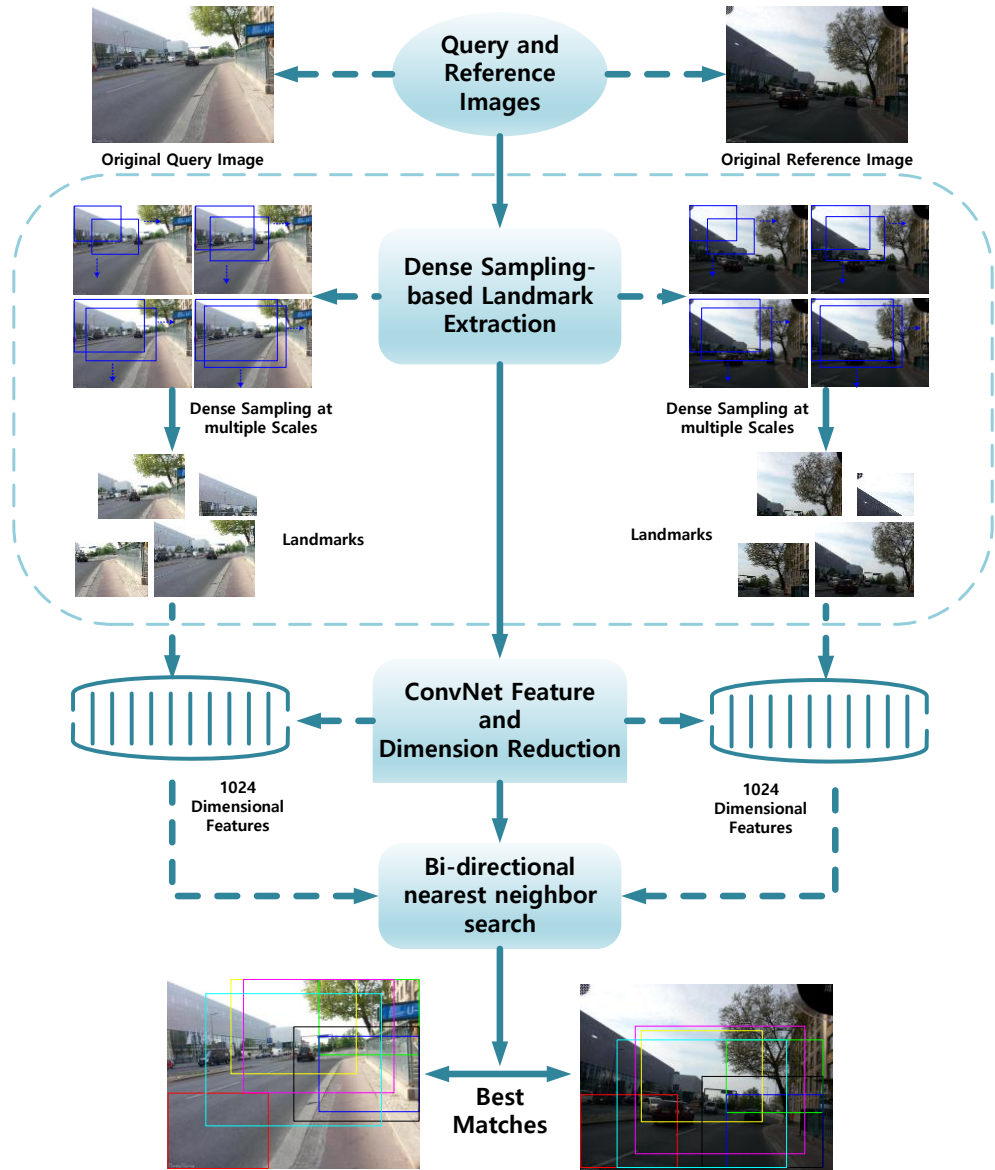

Fig. 8. Overview of our dense sampling based visual place recognition method. First, a total of 100 landmarks are extracted from different scale levels by dense sampling strategy. Next, 1,024 dimensional ConvNet features are computed for representing the landmarks before the bi-directional nearest neighbor search conducted to match pairwise landmarks. Thus, the best pairwise match is obtained by pooling all the pairwise matching results for similarity measure.

The overview of our system flowchart is summarized in Fig. 8. First, we leverage the dense sampling method for extracting a set of landmarks at multiple scale levels from both the query and the reference image. More specifically, a total number of 100 landmarks are obtained by the dense sampling imposed on the entire image along the image width and height with varying normalized scales and the sampling strides. In addition, the landmark is extracted maintaining the aspect ratio of original image.

An example of extracting landmarks from multiple scale levels by dense sampling is given in Fig. 9. It is shown that a total number of 100 landmarks are extracted from 4 different scale levels with respective 0.16, 0.25, 0.36, 0.49 normalized scales. Besides, 25 landmarks are extracted in each scale level from the whole image.

Thereafter, analogous to the classical method in [9], we compute the off-the-shelf CNN features of these local landmarks and perform bi-directional similarity matching between pairwise landmarks. The best pairwise match is obtained by pooling all the pairwise matching results, and thus used as the final similarity measure.

Compared to the classical method, the proposed method exhibits the following advantages: 1) our method allows extracting sufficient landmarks at multiple scales while somewhat maintains the uniform distribution of different scales and proportions of overlap between pairwise landmarks. 2) The landmark position achieved by our method can be pre-calculated instead of using Edge Box with desirable computational efficiency..

The parameters setting and performance of this proposed method for visual place recognition are given in the next section.

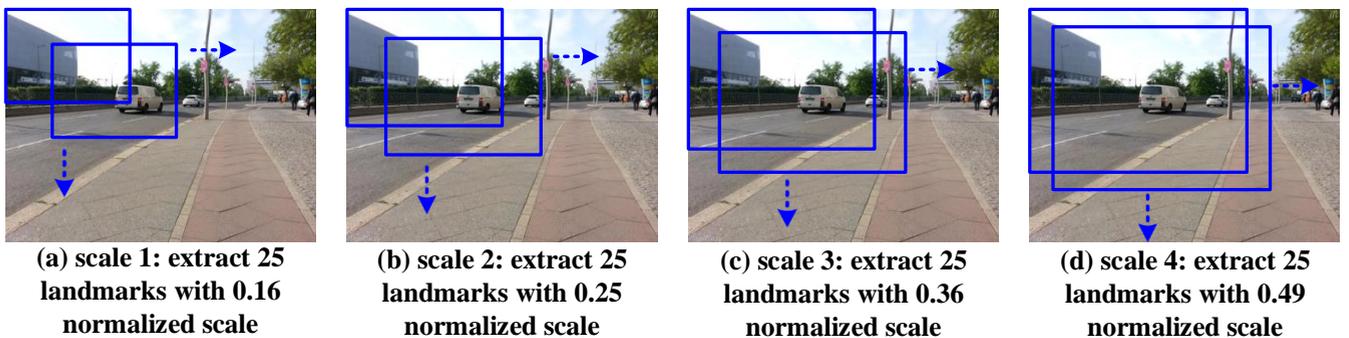

**(a) scale 1: extract 25 landmarks with 0.16 normalized scale**     **(b) scale 2: extract 25 landmarks with 0.25 normalized scale**     **(c) scale 3: extract 25 landmarks with 0.36 normalized scale**     **(d) scale 4: extract 25 landmarks with 0.49 normalized scale**

Fig. 9. An example of extracting landmarks from multiple scale levels by dense sampling. (a) scale 1, with 25 landmarks of 0.16 normalized scale extracted from the entire image, the landmark is cropped maintaining the aspect ratio of the original image ; (b) scale 2, the landmarks of 0.25 normalized scale are extracted in the same way as scale 1; (c) scale 3, the landmarks of 0.36 normalized scale are extracted in the same way as scale 1; (d) scale 4, the landmarks of 0.49 normalized scale are extracted in the same way as scale 1.

## 5 Experiments and results

In this section, we conduct extensive experiments to evaluate the performance of the proposed method on three challenging public benchmark datasets introduced in Section 2.

**5.1 Parameter selection for dense sampling method**

We carry out the parameter analysis for dense sampling method in the following three aspects: the number of scale levels, the number of landmarks per scale and the normalized scale of different scales. These parameters should be carefully determined based on the three observations in Section 3. Specifically, we have the following considerations in parameters selection:

1. The normalized scale of landmarks in different scale levels vary from 0.09 to 0.50 approximately. Generally, landmarks with smaller normalized scale have negative influence for place recognition while landmarks with larger normalized scale tends to take up the entire image which is lack of invariance for view point changes [15].
2. The appropriate distribution of landmarks should be guaranteed.
3. The proportion of overlap between pairwise landmarks per image should be considered.

According to these considerations, we conduct a comprehensive study for parameter selection. All the experiments in this subsection are conducted on the most challenging dataset, Mapillary Berlin. In addition, for the sake of consistency, the total number of extracted landmarks per image is set to be roughly 100.

Table 2 gives the specific experimental setting for appropriately choosing the number of scale levels. In these settings, we take into consideration both cases of single and multiple scale levels. In addition, we attempt to ensure the selected scales cover the range of normalized scales (0.09 to 0.5) available while maintain the discrimination between different scales. The corresponding results are shown in Fig. 10.

Table 2 Six different experimental settings with single and multiple scale levels

| Setting | | Normalized scale | Number Of landmarks | Total number |
|---|---|---|---|---|
| Single | Set 1 | 0.25 | 100 | 100 |
| | Set 2 | 0.49 | 100 | 100 |
| Multiple | Set 3 | 0.25 | 49 | 98 |
| | | 0.49 | 49 | |
| | Set 4 | 0.16 | 49 | 98 |
| | | 0.36 | 49 | |
| | Set 5 | 0.16 | 25 | 100 |
| | | 0.25 | 25 | |
| | | 0.36 | 25 | |
| | | 0.49 | 25 | |
| | Set 6 | 0.09 | 16 | 96 |
| | | 0.18 | 16 | |
| | | 0.25 | 16 | |
| | | 0.33 | 16 | |
| | | 0.41 | 16 | |
| | | 0.49 | 16 | |

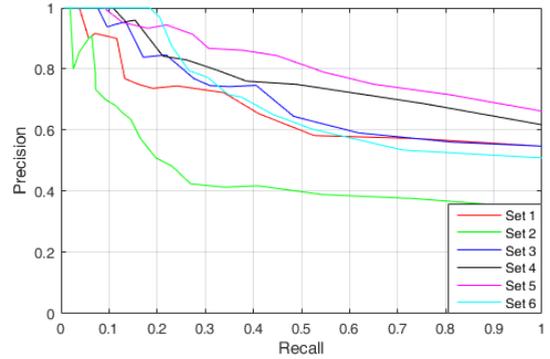

Fig. 10. Results of dense sampling method with different scale levels for place recognition on Mapillary Berlin dataset

It is observed that the setting based on single scale level achieve the worse performance, which indicates multiple scales are beneficial for improving the recognition accuracy. In particular, Set 2 achieves the worst performance resulting from the larger normalized scale landmarks lacking invariance of view point changes.

On the other hand, in terms of the settings involving multiple scales, Set 5 with four different scales reports the best performance and achieves superior results to the Set 3 and 4 with two different scales. However, Set 6 with six different scales achieves the worst performance which is even inferior to the single scale scenario. This can be explained by insufficient landmarks extracted per scale particularly at relatively small scales such that it is difficult to generate an appropriate proportion of overlap, and thus fails to provide a comprehensive description for whole image. Thus, sufficient landmarks per scale should be guaranteed for the recognition performance. In summary, the multi-scale works better than single scale, while the number of landmarks per scale should be sufficient. Due to the limitation of total number of landmarks in this paper, the number of scale levels is set to 4 for the tradeoff between the number of scales and the number of landmarks per scale.

Subsequently, we explore the influence of the number of landmarks per scale. In experimental setting, we use the four scale levels as 0.16, 0.25, 0.36 and 0.49 normalized scales and adopt five different numbers of landmarks extracted from per scale. The specific settings are shown in Table 3, while the experimental results are shown in Fig. 11 accordingly.

It is shown that the slight variance hardly produce any significant influence on recognition performance. Overall, Set 1 with uniform distribution of landmarks achieves the best performance while Set 4 with the most variation in landmark distribution obtains the worst performance, which indicates that uniform distribution leads to a descriptive representation of visual contents. Thus, the number of landmarks per scale is set to 25 in the remaining experiments.

Table 3 Five different experimental settings with fixed normalized scale and different number of landmarks per scale

| Normalized scale | Number Of landmarks per scale | | | | |
|---|---|---|---|---|---|
| | Set 1 | Set 2 | Set 3 | Set 4 | Set 5 |
| 0.16 | 25 | 36 | 36 | 49 | 49 |
| 0.25 | 25 | 25 | 36 | 25 | 36 |
| 0.36 | 25 | 25 | 16 | 16 | 9 |
| 0.49 | 25 | 16 | 16 | 9 | 9 |

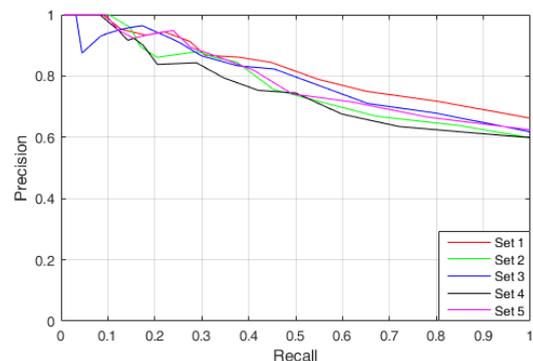

Fig. 11. Results of dense sampling method with fixed normalized scale and different number of landmarks per scale for place recognition on Mapillary Berlin dataset

Finally, we conduct four different groups of experiments to explore the influence of the normalized scales of different scales. We use the number of scales as 4, the number of landmarks per scale as 25 and four different normalized scale combinations for four scales. Specifically, Set 1 and Set 2 have similar setups, whilst Set 3 tends to use smaller size landmarks and oversize landmarks are utilized in Set 4. The specific settings are shown in Table 4 while the results are shown in Fig. 12.

Table 4 Four different experimental settings with fixed number of landmarks per scale and different normalized scales

| Number Of landmarks per scale | Normalized scale | | | |
| --- | --- | --- | --- | --- |
| | Set 1 | Set 2 | Set 3 | Set 4 |
| 25 | 0.16 | 0.18 | 0.09 | 0.20 |
| 25 | 0.25 | 0.27 | 0.20 | 0.32 |
| 25 | 0.36 | 0.38 | 0.32 | 0.44 |
| 25 | 0.49 | 0.51 | 0.44 | 0.54 |

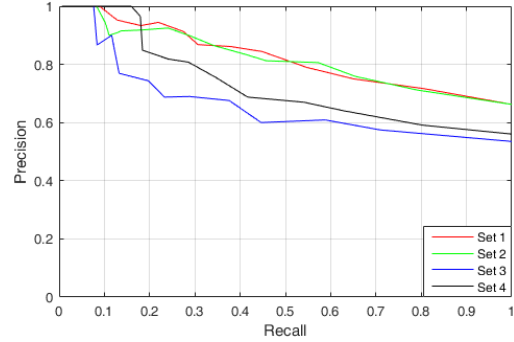

Fig. 12. Results of dense sampling method with fixed number of landmarks per scale and different normalized scales for place recognition on Mapillary Berlin dataset

In Fig. 12, Set 1 and Set 2 achieve the comparable performance while significantly outperform the other two settings. This implies either oversize or small-size landmarks adversely affects the recognition accuracy, since, the small-scale landmarks leads to unreliable matching while oversize landmarks lack the invariance of view point changes.

In addition, the performance of Set 1 and Set 2 are similar due to the similar setup. This implies the propose method enjoys the desirable invariance to slight scale changes. In this paper, the normalized scales of four different scales are set to 0.16, 0.25, 0.36, 0.49, respectively.

To sum up, although the landmarks with multiple scales benefit the place recognition, it is still necessary to balance the number of scale levels and the number of landmarks per scale. In addition, the uniform distribution and appropriate normalized scales of different scales also helps improving the performance of place recognition. Thus, we select four different scales with 0.16, 0.25, 0.36, 0.49 normalized scales, respectively to extract landmarks from whole image and we extract 25 landmarks for per scale in implementation.

In parameter selection, particularly, the normalized scale varies in the range of [0.15, 0.5] and superior results are reported when the normalized scale takes the value of 0.25 and 0.36. In addition, the number of landmarks per scale is carefully selected taking into account the uniform landmark distribution. Thus, the only tuning parameter in our case is the number of scale levels which is mainly based on the number of landmarks extracted per image. More specifically, the number of scale levels generally increases proportionally with the growing number of landmarks.

### 5.2 Comparative study

We compare our proposed visual place recognition method based on dense sampling with the landmark selection method proposed in Algorithm 1 for place recognition along with the classical method on three challenging dataset: 'day-left' vs 'night-right' sequences from Gardens Point Campus dataset, Mapillary Berlin dataset and 'summer 2015' vs 'winter 2012' sequences from Freiburg Across Seasons dataset. These three datasets include a wide variety of the challenging situations: large illumination and viewpoint variances, season and scenes changes due to dynamic objects and environment.

The experimental results of different methods on three datasets are shown in Fig. 13. It is observed that the performance of dense sampling method is slightly better than the landmark selection method, whereas significantly outperforms the classical method on all three datasets. Qualitatively, the proposed two methods utilize more reliable landmarks compared with the classical method. Particularly, the dense sampling method with suitable parameters allows more appropriate distribution of landmarks, and thus achieves the best performance. By contrast, the landmark selection method leads to uneven distribution of the landmarks yielding the suboptimal results.

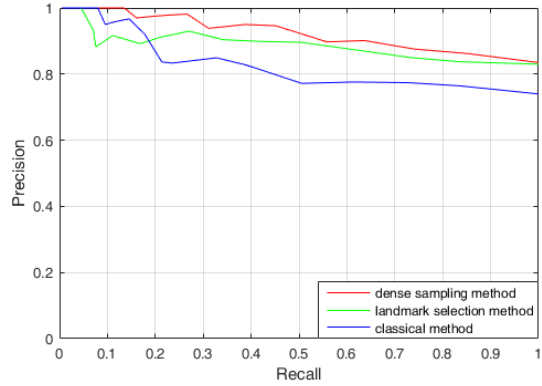

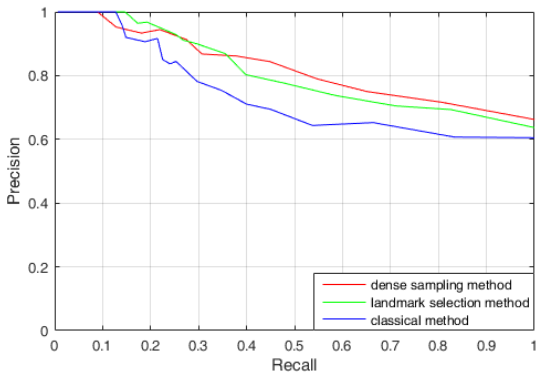
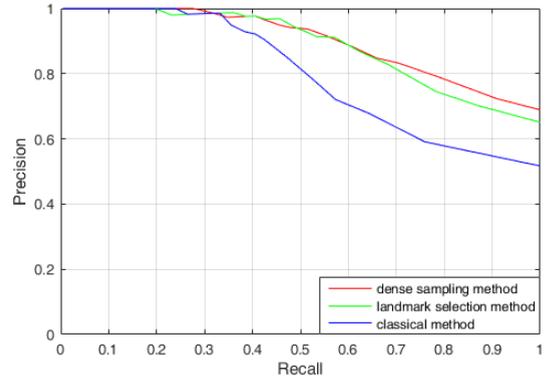

Fig. 13. (a) Results of dense sampling, landmark selection and classical method for place recognition on 'day-left' vs 'night-right' sequences from Gardens Point Campus dataset; (b) Results of dense sampling, landmark selection and classical method for place recognition on Mapillary Berlin dataset; (c) Results of dense sampling, landmark selection and classical method for place recognition on 'summer 2015' vs 'winter 2012' sequences from Freiburg Across Seasons dataset

On the other hand, the classical landmark-based visual place recognition method aims to leverage Edge Box method for extracting the representative object region, and thus tends to produce reliable matches to distinguish the query-dependent ground truth images and query-irrelevant images. However, top 100 candidate regions extracted by Edge Box hardly provide a comprehensive description for the entire image. Fig. 14 gives an example illustrating the coverage area of top 100 landmarks respectively extracted by Edge Box and dense sampling in the whole image. It is clearly shown that the landmarks resulting from Edge Box only cover about the half image and most landmarks are intensively distributed around the central region. Furthermore, relatively intensive distribution of landmarks extracted by Edge Box exists in most images and thus, it fails to represent the whole image comprehensively and prone to the perceptual aliasing problem leading to false matches. By contrast, in our proposed dense sampling based method, landmarks generated uniformly from the whole image allow providing a relatively comprehensive description of the original image. As presented in Fig. 14(b), the landmarks extracted by dense sampling method cover the whole image with extensive distribution. Thus, dense sampling based method outperforms the classical method in capturing the landmark distribution in an image.

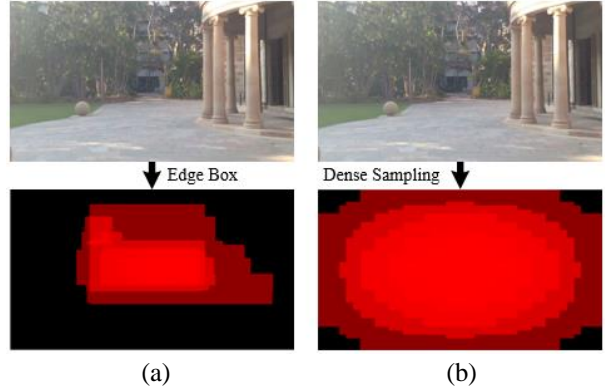

(a) (b)

Fig. 14. (a) An example of the coverage area of top 100 landmarks extracted by Edge Box in whole image. (b) An example of the coverage area of 100 landmarks extracted by dense sampling in whole image. Deeper color area indicates more coverage by landmarks. The image is from Gardens Point Campus dataset.

### 5.2 Analysis of the computational cost

We analyze the computational costs of the classical method and our methods quantitatively in Table 5. It is shown that the computational cost of the proposed method is lower than the classical method. To be specific, according to [16], with the same platform, landmark extraction by Edge Box costs about 0.25s which is similar to computing 100 CNN features of landmarks by AlexNet and far higher than the remaining steps between two images. However, since the position of landmarks can be pre-calculated in the proposed method, the computational complexity is about half of the classical method when matching two images.

In addition, for the landmark selection method, the cost of landmark extraction is difficult to determine, because it depends on specific scenes and images. Nevertheless, its computational cost is still higher than the classical method because it requires extracting more candidate regions and selecting suitable ones from them. On the other hand, we can produce a large set of candidate regions as many as 5000 or 10000 per image to select sufficient landmarks for satisfying the above observations. However, the computational cost of this strategy is unaffordable in practice.

To sum up, the proposed dense sampling method achieves the best performance with the desirable computational efficiency. By contrast, although the proposed landmark selection method works better than the classical method, it suffers from expensive computational cost, and thus is less feasible in real-world applications.

Table 5 Cost of classical method and dense sampling method when matching two images

|  | Extract landmarks | Computing CNN features | Remaining steps |
|---|---|---|---|
| Classical method | $\approx 0.25s$ | $\approx 0.21s$ | $< 0.01s$ |
| Dense sampling method | - | $\approx 0.21s$ | $< 0.01s$ |

## 6 Conclusions

In this paper, in order to address the limitations in classical landmark-based visual place recognition method, we carry out a comprehensive study to explore the impact of various landmark scales and proportion of overlap on the recognition performance and arrive at three beneficial observations accordingly. In a nutshell, characterizing local landmarks with proper scale and uniform distribution benefit the place recognition, while maintaining the appropriate proportion of overlap helps improving the descriptive power of the visual representation. Furthermore, inspired by the three observations, we propose a dense sampling based visual place recognition method. More specifically, this method uniformly generates the landmarks with multiple appropriate scales and certain overlap proportions from whole images. Thus, it provides a global representation for the entire image with sufficient descriptive power. The experimental results on three challenging datasets demonstrate that our method provides significant performance boost over the state-of-the-art method. Although the three observations result from an empirical study, we believe that they play a beneficial role in extracting and representing local landmarks for visual place recognition, and thus significantly contribute to improving the recognition performance. Therefore, they can serve as helpful implementation practices in real-world applications.

**Acknowledges**: This study is supported in part by the National Natural Science Foundation of China (Grant no. 61473085, 51775110, 61703096), the National Natural Science Foundation of Jiangsu Province (No.BK20170691), China Postdoctoral Science Foundation (No.2017M611655) and China Scholarship Council.